\documentclass[11pt,a4paper]{article}
\usepackage[acceptedWithA]{tacl2018v2}
\usepackage{times}
\usepackage{latexsym}

\usepackage{url}
\usepackage{xspace,mfirstuc,tabulary}

\usepackage{microtype}
\usepackage{graphicx}
\usepackage{subfigure}
\usepackage{booktabs} 
\usepackage{algorithm}
\usepackage{algorithmic}
\usepackage{times}
\usepackage{latexsym}
\usepackage{amsmath}
\usepackage{amssymb}
\usepackage{url}
\usepackage{bm}
\usepackage{natbib}
\usepackage{multirow}
\usepackage{graphicx}
\usepackage{caption}
\usepackage{subfigure}
\usepackage{soul}

\usepackage{hyperref}

\newcommand{\cQi}[1]{{\color{black}#1}}

\DeclareMathOperator*{\softmax}{\text{softmax}}

\title{Insertion-based Decoding with automatically \\Inferred Generation Order}

\def \nyu{$^\ddag$}
\def \ti{$^\diamond$}
\def \fair{$^\dagger$}

\author{
Jiatao Gu\fair, 
Qi Liu\ti\thanks{\; This work was completed while the author worked as an AI resident at Facebook AI Research.},
and
Kyunghyun Cho\nyu\fair\\
\fair Facebook AI Research \ \ti University of Oxford \\
\nyu New York University, CIFAR Azrieli Global Scholar  \\
\fair\texttt{\{jgu, kyunghyuncho\}@fb.com} \nyu\texttt{qi.liu@st-hughs.ox.ac.uk} \\
}

\date{}

\begin{document}
\maketitle
\begin{abstract}
  Conventional neural autoregressive decoding commonly assumes a fixed left-to-right generation order, which may be sub-optimal. In this work, we propose a novel decoding algorithm -- InDIGO -- which supports flexible sequence generation in arbitrary orders through insertion operations. We extend Transformer, a state-of-the-art sequence generation model, to efficiently implement the proposed approach, enabling it to be trained with either a pre-defined generation order or adaptive orders obtained from beam-search.
  Experiments on four real-world tasks, including word order recovery, machine translation, image caption and code generation, demonstrate that our algorithm can generate sequences following arbitrary orders, while achieving competitive or even better performance compared to the conventional left-to-right generation. The generated sequences show that InDIGO adopts adaptive generation orders based on input information.
\end{abstract}

\section{Introduction}
\label{intro}

Neural autoregressive models have become the \textit{de facto} standard in a wide range of sequence generation tasks, such as machine translation \cite{bahdanau2014neural}, summarization \cite{rush2015neural} and dialogue systems \cite{vinyals2015neural}. In these studies, a sequence is modeled autoregressively with the left-to-right generation order, which raises the question of whether generation in an arbitrary order is worth considering \cite{vinyals2015order,ford2018importance}. 
Nevertheless, previous studies on generation orders mostly resort to a fixed set of generation orders, showing particular choices of ordering are helpful~\cite{wu2018beyond,ford2018importance,NIPS2018_7796}, without providing an efficient algorithm for finding adaptive generation orders, or restrict the problem scope to $n$-gram segment generation~\cite{vinyals2015order}.
\begin{figure}[t]
    \centering
    \includegraphics[width=\linewidth]{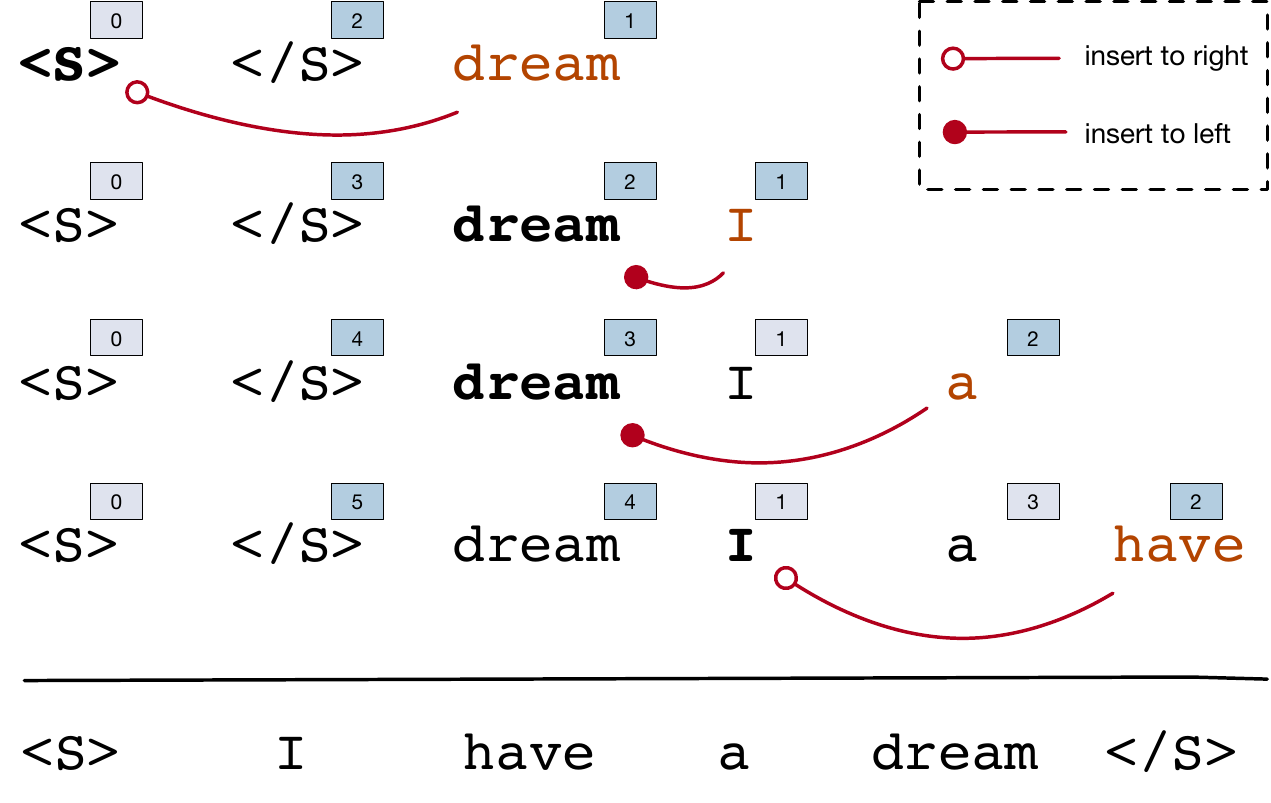}
    \caption{An example of InDIGO. At each step, we simultaneously predict the next token and its (relative) position to be inserted. 
    The final output sequence is obtained by mapping the words based on their positions.}
    \vspace{-7pt}
    \label{fig:insert_exp}
\end{figure}

In this paper, we propose a novel decoding algorithm, \textit{Insertion-based Decoding with Inferred Generation Order} (InDIGO), which models generation orders as latent variables and automatically infers the generation orders by simultaneously predicting a word and its position to be inserted at each decoding step.
Given that absolute positions are unknown before generating the whole sequence, we use a relative-position-based representation to capture generation orders. \cQi{
We show that decoding consists of a series of insertion operations with a demonstration shown in Figure~\ref{fig:insert_exp}.} 

We extend Transformer~\cite{vaswani2017attention} for supporting insertion operations, where the generation order is directly captured as relative positions through self-attention inspired by~\cite{shaw2018self}. 
For learning, we maximize the evidence lower-bound (ELBO) of the maximum likelihood objective, and study two approximate posterior distributions of generation orders based on a pre-defined generation order and adaptive orders obtained from beam-search, respectively.

Experimental results on word order recovery, machine translation, code generation and image caption demonstrate that our algorithm can generate sequences with arbitrary orders, while achieving competitive or even better performance compared to the conventional left-to-right generation. Case studies show that the proposed method adopts adaptive orders based on input information. The code will be released as part of the official repo of Fairseq (\url{https://github.com/pytorch/fairseq}).

\section{Neural Autoregressive Decoding}

Let us consider the problem of generating a sequence $\bm{y}=(y_1, ..., y_{T})$ conditioned on some inputs, e.g., a source sequence $\bm{x}=(x_1, ..., x_{T'})$. Our goal is to build a model \cQi{parameterized by} $\theta$ that models the conditional probability of $\bm{y}$ given $\bm{x}$, which is factorized as:
\begin{equation}
    p_\theta(\bm{y}|\bm{x}) = \prod_{t=0}^{T} p_\theta(y_{t+1} | y_{0:t}, x_{1:T'}),
    \label{eq.prob}
\end{equation}
where $y_0$ and $y_{T+1}$ are special tokens $\langle\text{s}\rangle$ and $\langle\text{/s}\rangle$, respectively. The model \cQi{sequentially predicts} the conditional probability of the next token \cQi{at} each step $t$, which can be implemented by any function approximator such as RNNs~\cite{bahdanau2014neural} and Transformer~\cite{vaswani2017attention}.

\paragraph{Learning}
Neural autoregressive model is commonly learned by maximizing the conditional likelihood $\log p(\bm{y}|\bm{x}) = \sum_{t=0}^T\log p_\theta(y_{t+1}|y_{0:t}, x_{1:T'})$ given a set of parallel examples.

\paragraph{Decoding}

A common way to decode a sequence from a trained model is to make use of the autoregressive nature that allows us to predict one word at each step. 
Given any source $\bm{x}$, we essentially follow the order of factorization to generate tokens sequentially using some heuristic-based algorithms such as greedy decoding and beam search. 

\section{Insertion-based Decoding with Inferred Generation Order (InDIGO)}
Eq.~\ref{eq.prob} explicitly assumes a \textit{left-to-right} (L2R) generation order of the sequence $\bm{y}$.
In principle, we can factorize the sequence probability in any permutation and train a model for each permutation separately. As long as we have infinite amount of data with proper optimization performed, all these models are equivalent. 
Nevertheless, \citet{vinyals2015order} have shown that the generation order of a sequence actually matters in many real-world tasks, e.g., language modeling. 

Although the L2R order is a strong inductive bias, as it is natural for most human-beings to read and write sequences from left to right, L2R is not \cQi{necessarily} the optimal option for generating sequences. For instance, people sometimes tend to think of central phrases first before building up a whole sentence; 
For programming languages, it is beneficial to be generated based on abstract syntax trees~\cite{yin2017syntactic}.

Therefore, a natural question arises, \textit{how can we decode a sequence in its best order?}

\subsection{Orders as Latent Variables}
\label{sec.latent}

We address this question by  modeling generation orders as latent variables. 
Similar to \citet{vinyals2015order}, we rewrite the target sequence $\bm{y}$ in a particular order $\bm{\pi}=(z_2, ..., z_T, z_{T+1}) \in \mathcal{P}_{T}$\footnote{
$\mathcal{P}_T$ is the set of all the permutations of $(1, ..., T)$.
} 
as a set $\bm{y}_{\bm{\pi}} = \{(y_2, z_2), ..., (y_{T+1}, z_{T+1})\}$, where $(y_t, z_t)$ represents the $t$-th generated token and its absolute position, respectively. 
Different from the common notation, the target sequence is $2$-step drifted because the two special tokens $(y_0,z_0) = (\langle\text{s}\rangle, 0)$ and $(y_1,z_1)=(\langle\text{/s}\rangle, T+1)$ are always prepended to represent the left and right boundaries, respectively.
Then, we model the conditional probability as the joint distribution of words and  positions by marginalizing all the orders: $$p_\theta(\bm{y}|\bm{x}) = \sum_{\bm{\pi} \in \mathcal{P}_T} p_\theta(\bm{y}_{\bm{\pi}}|\bm{x}),$$ where for each element:
\begin{equation}
    \begin{split}
        p_\theta(\bm{y}_{\bm{\pi}}|\bm{x}) &=
        p_\theta(\underline{y_{T+2}}|y_{0:T+1}, z_{0:T+1}, x_{1:T'}) \cdot 
        \\
        &
        \prod_{t=1}^{T} p_\theta(\underline{y_{t+1}, z_{t+1}}|y_{0:t}, z_{0:t}, x_{1:T'})
    \end{split}
    \label{eq.formulate}
\end{equation}
where the third special token $y_{T+2} = \langle\text{eod}\rangle$ is introduced to signal the end-of-decoding, and $p(y_{T+2}|\cdot)$ is the end-of-decoding probability.

At decoding time, the factorization allows us to decode autoregressively by predicting word $y_{t+1}$ and its position $z_{t+1}$ step by step. The generation order is \textit{automatically inferred} during decoding.


\subsection{Relative Representation of Positions}
\label{sec:implicit_pos}
It is difficult and inefficient to predict the absolute positions $z_t$ without knowing the actual length $T$. One solution is directly using the absolute positions $z_0^t, ..., z_t^t$ of the partial sequence $y_{0:t}$ at each autoregressive step $t$.
For example, the absolute positions for the sequence ($\langle\text{s}\rangle$, $\langle\text{/s}\rangle$, dream, I) are $(z_{0}^{t}=0, z_{1}^{t}=3, z_{2}^{t}=2, z_{3}^{t}=1)$ in Figure~\ref{fig:insert_exp} at step $t=3$. 
It is however inefficient to model such explicit positions using a single neural network without recomputing the hidden states for the entire partial sequence, as some positions are changed at every step (as shown in Figure~\ref{fig:insert_exp}).

\paragraph{Relative Positions}
We propose using relative-position representations $\bm{r}_{0:t}^t$ instead of absolute positions $z_{0:t}^t$.
We use a ternary vector $\bm{r}_i^t  \in \{-1, 0, 1\}^{t+1}$ as the relative-position representation for $z_i^{t}$.
The $j$-th element of $\bm{r}_i^t$ is defined as:
\begin{equation}
    \bm{r}_{i, j}^t = \begin{cases} 
    -1 & z_j^t > z_i^t \ \ (\text{left})\\ 
    0 & z_j^t = z_i^t \ \  (\text{middle})\\ 
    1 & z_j^t < z_i^t \ \ (\text{right})\end{cases},
    \label{eq.relative_pos}
\end{equation}
where the elements of $\bm{r}_i^t$ show the relative positions \cQi{with respect to} all the \cQi{other} words in the partial sequence at step $t$. 
We use a matrix $R^t = \left[\bm{r}_0^t, \bm{r}_1^t, ..., \bm{r}_t^t\right]$ to show the relative-position representations of all the words in the sequence.
The relative-position representation can always be mapped back to the absolute position $z^t_i$ by:
\begin{equation}
    z^t_i = \sum_{j=0}^t\max(0, \bm{r}_{i, j}^t)
    \label{eq.map_back}
\end{equation}
One of the biggest advantages for using such vector-based representations is that 
at each step, updating the relative-position representations is simply \textit{extending} the relative-position matrix $R^t$ with the next predicted relative position, because the (left, middle, right) relations described in Eq.~\eqref{eq.relative_pos} stay unchanged once they are created. Thus, we update $R^t$ as follows:
\begin{equation}
    R^{t+1} = \left[\begin{array}{ccc|c}
           &   &   &   \bm{r}_{t+1, 0}^{t+1}\\
           & R^t &   & \vdots \\
           &   &   &   \bm{r}_{t+1, t}^{t+1} \\
           \hline
          -\bm{r}_{t+1, 0}^{t+1}  & \cdots\ & -\bm{r}_{t+1, t}^{t+1} & 0 \\
\end{array}\right]
\label{eq.pos_update}
\end{equation}
where we use $\bm{r}_{t+1}^{t+1}$ to represent the  relative position at step $t+1$. This append-only property enables our method to reuse the previous hidden states without recomputing the hidden states at each step. For simplicity, the superscript of $\bm{r}$ is omitted from now on without causing conflicts.

\subsection{Insertion-based Decoding}

Given a partial sequence $y_{0:t}$ and its corresponding relative-position representations $\bm{r}_{0:t}$, not all of the $3^{t+2}$ possible vectors are valid for the next relative-position representation, $\bm{r}_{t+1}$. Only these vectors corresponding to \textit{insertion} operations satisfy Eq.~\eqref{eq.map_back}.  
In Algorithm~\ref{alg:insert_gen}, we describe an insertion-based decoding framework based on this observation. The next word $y_{t + 1}$ is predicted based on $y_{0:t}$ and $\bm{r}_{0:t}$. We then choose an existing word $y_k$ ($0 \leq k \leq t$)) from $y_{0:t}$ and insert $y_{t + 1}$ to its left or right. As a result, the next position $\bm{r}_{t+1}$ is determined by
\begin{equation}
    \bm{r}_{t+1, j} = \begin{cases} 
    s & j = k  \\ 
    \bm{r}_{k, j} & j \neq k
    \end{cases}, \ \ \ \  \forall j \in [0, t]
    \label{eq.insertion}
\end{equation}
where $s = -1$ if $y_{t + 1}$ is on the left of $y_{k}$, and $s = 1$ otherwise. Finally, we use $\bm{r}_{t+1}$ to update the relative-position matrix $R$ as shown in Eq.~\eqref{eq.pos_update}.

\begin{algorithm}[tb]
  \caption{Insertion-based Decoding}
  \small
  \label{alg:insert_gen}
\begin{algorithmic}
  \STATE {\bfseries Initialize:} $\bm{y}=\left(\langle\text{s}\rangle, \langle\text{/s}\rangle\right)$, $R = 
            \scriptsize{\begin{bmatrix}
            0  & 1 \\
            -1 & 0  
            \end{bmatrix}}$, $t=1$
  \REPEAT
  \STATE {Predict the next word $y_{t+1}$ based on $\bm{y}, R$}.
  \IF{$y_{t+1}$ is $\langle\text{eod}\rangle$}
  \STATE {break}
  \ENDIF
  \STATE {Choose an existing word $y_k \in \bm{y}$;}
  \STATE {Choose the left or right ($s$) of $y_k$ to insert;}
  \STATE {Obtain the next position $\bm{r}_{t+1}$ with $k, s$ (Eq.~\eqref{eq.insertion}).}
  \STATE {Update $R$ by appending $\bm{r}_{t+1}$ (Eq.~\eqref{eq.pos_update}).}
  \STATE {Update $\bm{y}$ by appending $y_{t+1}$}
  \STATE {Update $t = t + 1$}
  \UNTIL{Reach the maximum length}
  \STATE{Map back to absolute positions $\bm{\pi}$ (Eq.~\eqref{eq.map_back})}
  \STATE{Reorder $\bm{y}$: $y_{z_i} = y_i \ \ \ \forall z_i \in \bm{\pi}, i \in [0, t]$ }
\end{algorithmic}
\end{algorithm}


\section{Model}
\label{sec.insertable}
\begin{figure*}[htbp]
    \centering
    \includegraphics[width=\linewidth]{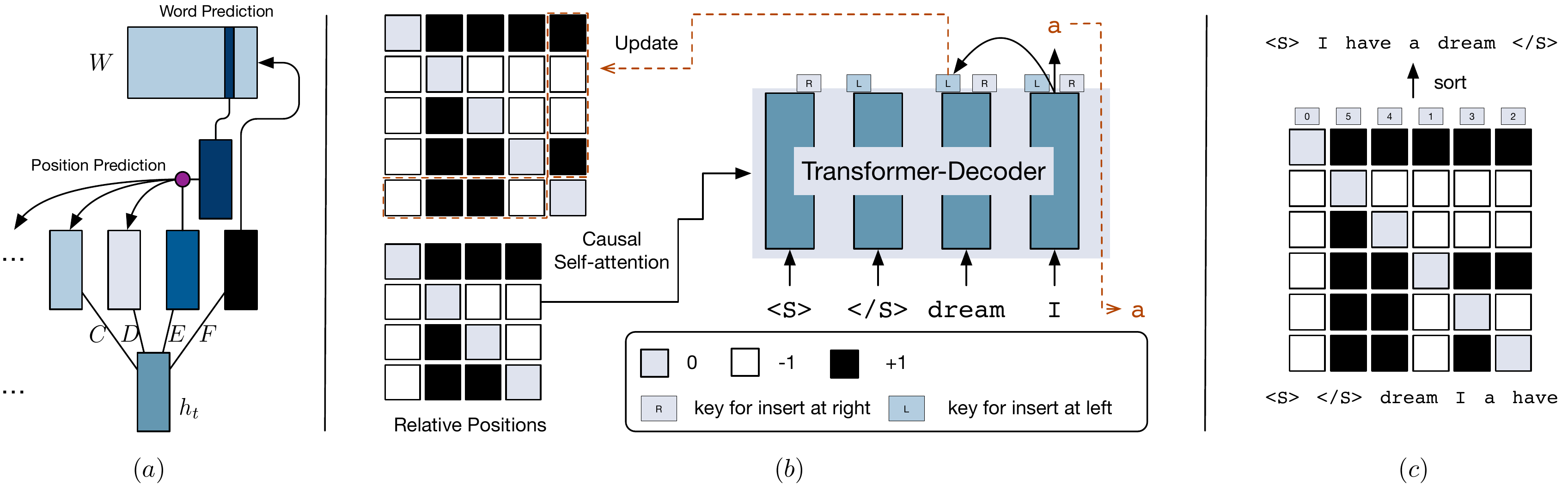}
    \caption{The overall framework of the proposed Transformer-InDIGO which includes (a) the word \& position prediction module; (b) the one step decoding with position updating; (c) final decoding output by reordering. The black-white blocks represent the relative position matrix.}
    \label{fig:framework}
\end{figure*}

We present Transformer-InDIGO, an extension of Transformer~\cite{vaswani2017attention}, supporting insertion-based decoding.
The overall framework is shown in Figure~\ref{fig:framework}.


\subsection{Network Design}
We extend the decoder of Transformer with relative-position-based self-attention, joint word \& position prediction and position updating modules.

\paragraph{Self-Attention}

One of the major challenges that prevents the vanilla Transformer from generating sequences following arbitrary orders is that the absolute-position-based positional encodings are inefficient as mentioned in Section \ref{sec:implicit_pos}, in that absolute positions are changed during decoding, invalidating the previous hidden states. In contrast, we adapt \newcite{shaw2018self} to use relative positions in self-attention. Different from \citet{shaw2018self}, in which a clipping distance $d$ (usually $d \geq 2$) is set for relative positions, our relative-position representations only preserve $d=1$ relations (Eq.~\eqref{eq.relative_pos}). 

Each attention head in a multi-head self-attention module of Transformer-InDIGO takes the hidden states of a partial sequence $y_{0:t}$, denoted as $U = (\bm{u}_0, . . . , \bm{u}_t)$, and its corresponding relative position matrix $R^t$ as input, where each input state $\bm{u}_i \in \mathbb{R}^{d_\text{model}}$. The logit $e_{i, j}$ for attention is computed as:
\begin{equation}
    e_{i, j} = \frac{\left(\bm{u}_i^\top Q\right) \cdot \left(\bm{u}_j^\top K + A_{[\bm{r}_{i, j} +1]}\right)^\top}{\sqrt{d_\text{model}}},
\end{equation}
where $Q, K \in \mathbb{R}^{d_\text{model} \times d_\text{model}}$ and $A \in \mathbb{R}^{3 \times d_\text{model}}$ are parameter matrices. \cQi{$A_{[\bm{r}_{i, j} + 1]}$} is the row vector indexed by $\bm{r}_{i, j} + 1$, which biases all the input keys based on the relative position, $\bm{r}_{i, j}$.  


\paragraph{Word \& Position Prediction}

Like the vanilla Transformer, we take the representations from the last layer of self-attention, $H = \left(\bm{h}_0, ... ,\bm{h}_t\right)$ and $H \in \mathbb{R}^{d_\text{model}\times (t+1)}$, to predict both the next word $y_{{t+1}}$ and its position vector $\bm{r}_{t+1}$ in two stages based on the following factorization:
$$p(y_{{t+1}}, \bm{r}_{t+1} | H)\! = \! p(y_{{t+1}}| H) \cdot p(\bm{r}_{t+1}|y_{{t+1}}, H)$$ 
It can also be factorized as predicting the position before predicting the next word, yet our preliminary experiments show that predicting the word first works slightly better. The prediction module for word \& position prediction are shown in Figure~\ref{fig:framework}(a). 



First, we predict the next word $y_{{t+1}}$ from the categorical distribution $p_\text{word}(y|H)$ as:
\begin{equation}
    p_\text{word}(y|H) = \softmax\left((\bm{h}_t^\top F) \cdot W^\top \right),
\end{equation}
where $W \in \mathbb{R}^{d_\text{V} \times d_\text{model}}$ is the embedding matrix and $d_\text{V}$ is the size of vocabulary. We linearly project the last representation $\bm{h}_t$ using $F \in \mathbb{R}^{d_\text{model} \times d_\text{model}}$ for querying $W$.
Then, as shown in Eq.~\eqref{eq.insertion}, the prediction of the next position is done by performing insertion operations to existing words which can be modeled similarly to Pointer Networks~\cite{vinyals2015pointer}. We predict a pointer $k_{t+1}\in [0, 2t+1]$ based on:
    \begin{equation}
        \begin{split}
         &p_{\text{pointer}}(k|y_{{t+1}}, H) =  \\
         &\softmax \left( 
         (\bm{h}_t^\top E  + W_{[y_{{t+1}}]}) 
         \cdot 
         {\begin{bmatrix}H^\top C\\ H^\top D\end{bmatrix}}^\top 
         \right),
        \end{split}
        \label{eq.pointer}
    \end{equation}
    where $C, D, E \in \mathbb{R}^{d_\text{model} \times d_\text{model}}$ and $W_{[y_{t+1}]}$ is the embedding of the predicted word. $C, D$ are used to obtain the left and right keys, respectively, considering that each word has two ``keys'' (its left and right) for inserting the generated word. 
    The query vector is obtained by adding up the word embedding $W_{[y_{t+1}]}$, and the linearly projected state, $\bm{h}_t^\top E$. 
    The resulting relative-position vector, $\bm{r}_{t+1}$ is computed using $k_{t+1}$ according to Eq.~\eqref{eq.insertion}. We manually set $p_{\text{pointer}}(0|\cdot) = p_{\text{pointer}}(2 + t|\cdot) = 0$ to avoid any word from being inserted to the left of $\langle\text{s}\rangle$ and the right of $\langle\text{/s}\rangle$.

\begin{table*}[t]
	\centering
	\footnotesize
	\scalebox{0.98}{
	\begin{tabular}{l | l}
	\toprule
	Pre-defined Order & Descriptions \\
	\midrule 
    Left-to-right (L2R)  & Generate words from left to right. \cite{wu2018beyond} \\
    Right-to-left (R2L)  & Generate words from right to left. \cite{wu2018beyond} \\
    \midrule
	Odd-Even (ODD)  &  Generate words at odd positions from left to right, then generate even positions. \cite{ford2018importance}\\
	Balanced-tree (BLT) & Generate words with a top-down left-to-right order from a balanced binary tree. \cite{stern2019insertion} \\ 
	Syntax-tree (SYN) & Generate words with a top-down left-to-right order from the dependency tree. \cite{wang2018tree}\\
	Common-First (CF) & Generate all common words first from left to right, and then generate the others. \cite{ford2018importance}\\
	Rare-First (RF) & Generate all rare words first from left to right, and then generate the remaining.  \cite{ford2018importance}\\
	\midrule
	Random (RND) & Generate words in a random order shuffled every time the example was loaded. \\
	\bottomrule
    	\end{tabular}}
	\caption{Descriptions of the pre-defined orders used in this work. Major references that have explored these generation orders with different models and applications are also marked.} 
	\vspace{0pt}
	\label{table:D2}
\end{table*}

\paragraph{Position Updating} 

As mentioned in Sec.~\ref{sec.latent}, we  update the relative position representation $R^t$ with the predicted $r_{t+1}$.
Because updating the relative positions will not change the pre-computed relative-position representations, Transformer-InDIGO can reuse the previous hidden states in the next decoding step the same as the vanilla Transformer.


\subsection{Learning}
Training requires maximizing the marginalized likelihood in Eq.~\eqref{eq.formulate}. 
Yet this is intractable since we need to enumerate all of the $T!$ permutations of tokens.
Instead, we maximize the evidence lower-bound (ELBO) of the original objective by introducing an approximate posterior distribution
of generation orders $q(\bm{\pi}|\bm{x}, \bm{y})$, which provides the probabilities of latent generation orders based on the ground-truth sequences $\bm{x}$ and $\bm{y}$:
\begin{equation}
\label{eq.variational}
\begin{aligned}
&\mathcal{L}_{\text{ELBO}} 
=\mathop{\mathbb{E}}_{\bm{\pi} \sim q} \log p_\theta(\bm{y}_{\bm{\pi}} | \bm{x}) + \mathcal{H}(q)\\ 
&=\mathop{\mathbb{E}}_{\bm{r}_{2:T+1} \sim q}
\left({
\sum_{t=1}^{T+1}\underbrace{\log p_\theta(y_{{t+1}}| y_{{0:t}}, \bm{r}_{0:t}, x_{1:T'})}_{\text{Word Prediction Loss}} }\right.\\
& + \left.{ 
\sum_{t=1}^{T}\underbrace{\log p_\theta(\bm{r}_{t+1}| y_{{0:t+1}}, \bm{r}_{0:t}, x_{1:T'})}_{\text{Position Prediction Loss}} }\right) + \mathcal{H}(q),
\end{aligned}
\end{equation}
where $\bm{\pi} = \bm{r}_{2:T+1}$, sampled from $q(\pi|\bm{x}, \bm{y})$, is represented as relative positions. $\mathcal{H}(q)$ is the entropy term which can be ignored if $q$ is fixed during training.
Eq.~\eqref{eq.variational} shows that given a sampled order, the learning objective is divided into word \& position objectives. For calculating the position prediction loss, we aggregate the two probabilities corresponding to the same position by
\begin{equation}
    p_\theta(\bm{r}_{t+1}|\cdot) = p_{\text{pointer}}(k^l|\cdot) + p_{\text{pointer}}(k^r|\cdot),  
\end{equation}
where $p_{\text{pointer}}(k^l|\cdot)$ and $p_{\text{pointer}}(k^r|\cdot)$ are calculated simultaneously from the same softmax function in Eq.~\eqref{eq.pointer}.
$k^l, k^r (k^l \neq k^r)$ represent the keys corresponding to the same relative position.

Here, we study two types of $q(\bm{\pi}|\bm{x}, \bm{y})$:

\paragraph{Pre-defined Order}
If we already possess some prior knowledge about the sequence, e.g., the L2R order is proven to be a strong baseline in many scenarios, we assume a Dirac-delta distribution $q(\bm{\pi}|\bm{x}, \bm{y}) = \delta(\bm{\pi}=\bm{\pi}^*(\bm{x}, \bm{y}))$,
where $\bm{\pi}^*(\bm{x}, \bm{y}))$ is a predefined order. 
In this work, we study a set of pre-defined orders which can be found in Table.~\ref{table:D2}, for evaluating their effect on generation.

\paragraph{Searched Adaptive Order (SAO)}

We choose the approximate posterior $q$ as the point estimation that maximizes $\log p_\theta(\bm{y}_{\bm{\pi}}|\bm{x})$, which can also be seen as the maximum-a-posteriori (MAP) estimation on the latent order $\bm{\pi}$.
In practice, we approximate these generation orders $\pi$ through \textit{beam-search}~\citep{pal2006sparse}. Unlike the original beam-search for autoregressive decoding that searches in the sequence space to find the sequence maximizing the probability shown in Eq. \ref{eq.prob}, we search in the space of all the permutations of the target sequence to find $\pi$ maximising Eq. \ref{eq.formulate}, as all the target tokens are known in advance during training.

More specifically, 
we maintain $B$ sub-sequences with the maximum probabilities using a set $\mathcal{B}$ at each step $t$. For every sub-sequence $y^{(b)}_{{0:t}} \in \mathcal{B}$, we evaluate the probabilities of every possible choice from the remaining words $y' \in \bm{y} \setminus y^{(b)}_{{0:t}}$ and its position $\bm{r}'$. We calculate the cumulative likelihood for each $y', \bm{r}'$,
based on which we select top-$B$ sub-sequences as the new set $\mathcal{B}$ for the next step.
After obtaining the $B$ generation orders, we optimize our objective as an average over these orders:
\begin{equation}
    \mathcal{L}_{SAO} = \frac{1}{B}\sum_{\pi \in \mathcal{B}}\log p_\theta(\bm{y}_{\bm{\pi}} | \bm{x})
 \end{equation}
where we assume $q(\bm{\pi}|\bm{x}, \bm{y})  = {\small \begin{cases} 
    1/B &  \bm{\pi} \in \mathcal{B}\\ 
    0 & \text{otherwise}\end{cases}}$.


\paragraph{Beam Search with Dropout}
The goal of beam search is to approximately find the most likely generation orders, which limits learning from exploring other generation orders that may not be favourable currently but may ultimately be deemed better.
Prior research~\cite{vijayakumar2016diverse} also pointed out that the search space of the standard beam-search is restricted. 
We encourage exploration by injecting noise during beam search \cite{cho2016noisy}. Particularly, we found it effective to keep the dropout on (e.g., dropout $=0.1$). 

\paragraph{Bootstrapping from a Pre-defined Order}

During preliminary experiments, sequences returned by beam-search were often degenerated by always predicting common or functional words (e.g., ``the'', ``,'', etc.) as the first several tokens, leading to inferior performance.
We conjecture that is due to the fact that the position prediction module learns much faster than the word prediction module, and it quickly captures spurious correlations induced by a poorly initialized model.
It is essential to balance the learning progress of these modules. To do so, we bootstrap learning  by pre-training the model with a pre-defined order (e.g., L2R), before training with beam-searched orders.

\subsection{Decoding}
\label{sec.decoding}
As for decoding,
we directly follow Algorithm~\ref{alg:insert_gen} to sample or decode greedily from the proposed model. However, in practice beam-search is important to explore the output space for neural autoregressive models. In our implementation, we perform beam-search for InDIGO as a two-step search. Suppose the beam size $B$, at each step, we do beam-search for word prediction and then with the searched words, try out all possible positions and select the top-$B$ sub-sequences. In preliminary experiments, we also tried doing beam-search for word and positions simultaneously with their joint probability. However, it did not seem helpful.

\section{Experiments}
We evaluate InDIGO extensively on four challenging sequence generation tasks: word order recovery, machine translation, natural language to code generation~\citep[NL2Code,][]{ling2016latent} and image captioning. We compare our model trained with the pre-defined orders and the adaptive orders obtained by beam-search. We use the same architecture for all orders including the standard L2R order. 

\subsection{Experimental Settings}

\paragraph{Dataset} 

The machine translation experiments are conducted on three language pairs for studying how the decoding order influences the translation quality of languages with diversified characteristics: WMT'16 Romanian-English (Ro-En),\footnote{
\url{http://www.statmt.org/wmt16/translation-task.html}
} WMT 18 English-Turkish (En-Tr)\footnote{
\url{http://www.statmt.org/wmt18/translation-task.html}
} 
and KFTT English-Japanese~\citep[En-Ja,][]{neubig11kftt}~\footnote{\url{http://www.phontron.com/kftt/}.} 
The English part of the Ro-En dataset is used for the word order recovery task.
For the NL2Code task, We use the Django dataset~\cite{oda2015ase:pseudogen1}\footnote{
\url{https://github.com/odashi/ase15-django-dataset}
} 
and the MS COCO~\cite{lin2014microsoft} with the standard split~\cite{karpathy2015deep} for the NL2Code task and image captioning, respectively. 
The dataset statistics are shown in Table~\ref{table:D1}. 

\begin{table}[t]
	\centering
	\small
	\scalebox{1.00}{
	\begin{tabular}{l | rrrr}
	\toprule
	Dataset & Train & Dev & Test & Length \\
	\midrule
    WMT16 Ro-En  & 620k & 2000 & 2000 & 26.48 \\
	WMT18 En-Tr  & 207k & 3007 & 3000 & 25.81 \\
    KFTT En-Ja   & 405k & 1166 & 1160 & 27.51 \\
    Django  & 16k  & 1000 & 1801 & 8.87 \\
    MS-COCO & 567k & 5000 & 5000 & 12.52 \\
	\bottomrule
	\end{tabular}}
	\caption{Dataset statistics for the machine translation, code generation and image captioning tasks. Length represents the average number of tokens for target sentences of the training set.} 
	\label{table:D1}
\end{table}


\paragraph{Preprocessing} 

We apply the Moses tokenization\footnote{
\url{https://github.com/moses-smt/mosesdecoder}\label{ref:moses}
} 
and normalization on all the text datasets except for codes. We perform $32,000$ joint BPE~\cite{sennrich2015neural} operations for the MT datasets, while using all the unique words as the vocabulary for NL2Code. For image captioning, we follow the same procedure as described by \newcite{lee2018deterministic}, where we use $49$ $512$-dimensional image feature vectors (extracted from a pretrained ResNet-18~\cite{he2016deep}) as the input to the Transformer encoder. The image features are fixed during training.

\begin{table*}[t]
	\centering
	\small
	\scalebox{0.9}{
	\begin{tabular}{l cccc  cccc  cccc }
	\toprule
    \multirow{2}{*}{Order} & \multicolumn{4}{c}{WMT16 Ro $\rightarrow$ En} & \multicolumn{4}{c}{WMT18 En $\rightarrow$ Tr} & \multicolumn{4}{c}{KFTT En $\rightarrow$ Ja} \\
    & BLEU & Ribes & Meteor & TER & BLEU & Ribes & Meteor & TER  & BLEU & Ribes & Meteor & TER \\
    \midrule
    RND & 20.20 & 79.35& 41.00& 63.20 &03.04& 55.45 & 19.12 & 90.60 & 17.09 & 70.89 & 35.24 & 70.11 \\
    \midrule 
	L2R & 31.82 & 83.37 & 52.19 & 50.62 & 14.85 & 69.20 & 33.90 & \textbf{71.56} & 30.87 & 77.72 & 48.57 & 59.92 \\
	R2L & 31.62 & 83.18 & 52.09 & 50.20 & 14.38 & 68.87 & 33.33 & 71.91 & 30.44 & \textbf{77.95} & 47.91 & 61.09 \\
	ODD & 30.11 & 83.09 & 50.68 & 50.79 & 13.64 & 68.85 & 32.48 & 72.84 & 28.59 & 77.01 & 46.28 & 60.12  \\
	BLT & 24.38 & 81.70 & 45.67 & 55.38 & 08.72 & 65.70 & 27.40 & 77.76 & 21.50 & 73.97 & 40.23 & 64.39 \\
	SYN & 29.62 & 82.65 & 50.25 & 52.14 & \multicolumn{4}{c}{--} & \multicolumn{4}{c}{--}\\
	CF  & 30.25 & 83.22 & 50.71 & 50.72 & 12.04 & 67.61 & 31.18 & 74.75 & 28.91 & 77.06 & 46.46 & 61.56 \\
	RF  & 30.23 & 83.29 & 50.72 & 51.73 & 12.10 & 67.44 & 30.72 & 73.40 & 27.35 & 76.40 & 45.15 & 62.14\\
	\midrule
	SAO  & \textbf{32.47} & \textbf{84.10} & \textbf{53.00} & \textbf{49.02} & 
	\textbf{15.18} & \textbf{70.06} & \textbf{34.60} & \textbf{71.56} & \textbf{31.91} & 77.56 & \textbf{49.66} & \textbf{59.80} \\
	\bottomrule
	\end{tabular}}
	\caption{Results of translation experiments for three language pairs in different decoding orders. Scores are reported on the test set with four widely used evaluation metrics (BLEU$\uparrow$, Meteor$\uparrow$, TER$\downarrow$ and Ribes$\uparrow$). We do not report models trained with SYN order on En-Tr and En-Ja due to the lack of reliable dependency parsers. The statistical significance analysis\textsuperscript{{\ref{ref:moses}}} between the outputs of SAO and L2R are conducted using BLEU score as the metric, and the p-values are $\leq 0.001$ for all three language pairs.} 
	\label{table:D3}
\end{table*}

\paragraph{Models} 

We set $d_\text{model}=512$, $d_\text{hidden}=2048$, $n_\text{heads}=8$, $n_\text{layers}=6$, $\text{lr}_{\max}=0.0005$, $\text{warmup}=4000$ and $\text{dropout}=0.1$ throughout all the experiments. The source and target embedding matrices are shared except for En-Ja, as our preliminary experiments showed that keeping the embeddings not shared significantly improves the translation quality.
Both the encoder and decoder use relative positions during self-attention except for the word order recovery experiments (where the position embedding is removed in the encoder, as there is no ground-truth position information in the input.)
We do not introduce task-specific modules such as copying mechanism~\cite{gu2016incorporating}.
\begin{figure}[t]
    \centering
    \includegraphics[width=\linewidth]{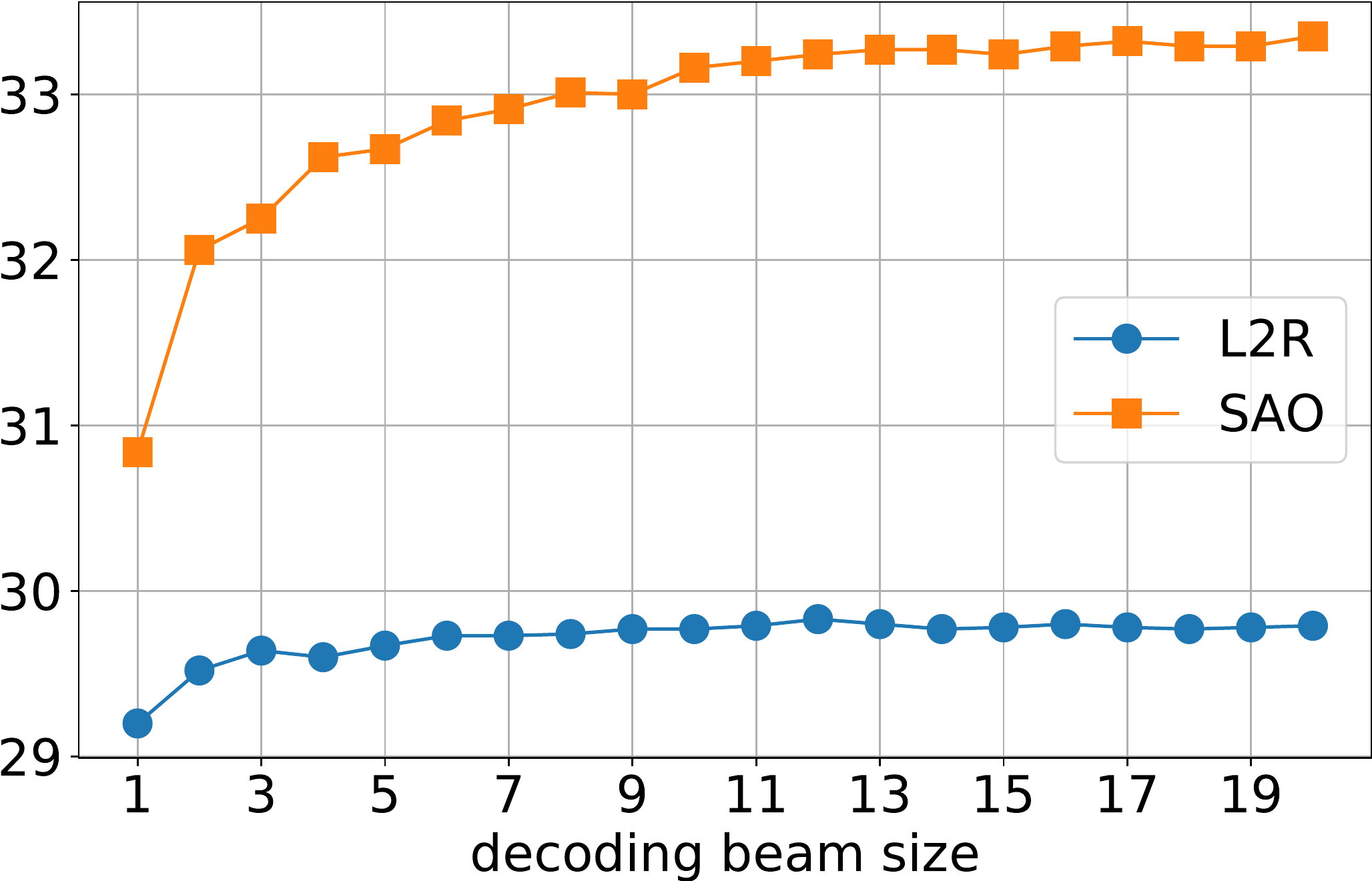}
    \caption{The BLEU scores on the test set for word order recovery with various decoding beam sizes.}
    \label{fig:reordering}

\end{figure}

\paragraph{Training}

When training with the pre-defined orders, we reorder words of each training sequence in advance accordingly which provides supervision of the ground-truth positions that each word should be inserted. 
We test the pre-defined orders listed in Table~\ref{table:D2}. 
The SYN orders were generated according to the dependency parse obtained by a dependency parse parser from Spacy~\cite{spacy2}
following a parent-to-children left-to-right order. 
The CF \& RF orders are obtained based on vocabulary cut-off so that the number of common words and the number of rare words are approximately the same~\cite{ford2018importance}.
We also consider on-the-fly sampling a random order for each sentence as the baseline (RND).
When using L2R as the pre-defined order, Transformer-InDIGO is almost equivalent to the vanilla Transformer, as the position prediction simply learns to predict the next position as the left of the $\langle\text{s}\rangle$ symbol. The only difference is that it enhances the vanilla Transformer with a small number of additional parameters for the position prediction. 

We also train Transformer-InDIGO using the searched adaptive order (SAO) where we set the beam size to $8$. In default, models trained with SAO are bootstrapped from a slightly pre-trained (6,000 steps) model in L2R order.

\paragraph{Inference}
During the test time, we do beam-search as described in Sec.~\ref{sec.decoding}. We observe from our preliminary experiments that models trained with different orders (either pre-defined or SAO) have very different optimal beam sizes for decoding. Therefore, we perform sensitivity studies, in which the beam sizes vary from $1\sim 20$ and pick the beam size with the highest BLEU score on the validation set for each particular model.

\subsection{Results and Analysis}
\begin{table}[t]
	\centering
	\small
	\scalebox{1.0}{
	\begin{tabular}{l  cc cc}
	\toprule
	\multirow{2}{*}{Model} & \multicolumn{2}{c}{Django} & \multicolumn{2}{c}{MS-COCO} \\
	& BLEU & Accuracy& BLEU & CIDEr-D \\
	\midrule
	L2R & 36.74 & 13.6\% & 22.12 & 68.88\\
	SAO & \textbf{42.33} & \textbf{16.3\%} & \textbf{22.58} & \textbf{69.42} \\
	\bottomrule
    \end{tabular}}
    \caption{Results on the official test sets for both code generation and image captioning tasks.}
    \label{table:D4}
    
\end{table}

\begin{table}[t]
    \centering
    \begin{tabular}{lcc}
    \toprule
    Model Variants & dev & test \\
     \midrule
    Baseline L2R & 32.53  & 31.82 \\
    SAO default                  & \textbf{33.60} & \textbf{32.47} \\
    \; \; no bootstrap           & 32.86 & 31.88 \\
	\; \; no bootstrap, no noise & 32.64 & 31.72 \\
	\; \; bootstrap from R2L order & 33.12 & 32.02 \\
	\; \; bootstrap from SYN order & 33.09 & 31.93 \\
	 \midrule
	\newcite{stern2019insertion} - Uniform & 29.99 & 28.52 \\
    \newcite{stern2019insertion} - Binary & 32.27 & 30.66 \\
    \bottomrule
    \end{tabular}
    \caption{Ablation study for machine translation on WMT16 Ro-En. Results of \newcite{stern2019insertion} are based on greedy decoding with the EOS penalty.}
    \label{table:ABS}
\end{table}
\paragraph{Word Order Recovery}

Word order recovery takes a bag of words as input and recovers its original word order, which is challenging as the search space is factorial. We do not restrict the vocabulary of the input words.
We compare our model trained with the L2R order and eight searched adaptive orders (SAO) from beam search for word order recovery. The BLEU scores over various  beam sizes are shown in Figure~\ref{fig:reordering}. 
The model trained with SAO lead to higher BLEU scores over that trained with L2R
with a gain up to $3$ BLEU scores. Furthermore, increasing the beam size brings more improvements for SAO compared to L2R, suggesting that InDIGO produces more diversified predictions so that it has higher chances to recover the order.


\paragraph{Machine Translation}


As shown in Table~\ref{table:D3}, we compare our model trained with pre-defined orders and the searched adaptive orders (SAO) with varying setups. 
We use four evaluation metrics including BLEU~\cite{papineni2002bleu}, Ribes~\cite{isozaki2010automatic}, Meteor~\cite{banerjee2005meteor} and TER~\cite{snover2006study} to avoid using a single metric that might be in favor of a particular generation order.  
Most of the pre-defined orders (except for the random order and the balanced tree (BLT) order) perform reasonably well with InDIGO on the three language pairs. The best score with a predefined word ordering is reached by the L2R order among the pre-defined orders except for En-Ja, where the R2L order works slightly better according to Ribes. 
This indicates that in machine translation, the monotonic orders are reasonable and reflect the languages. 
ODD, CF and RF show similar performance, which is below the L2R and R2L orders by around $2$ BLEU scores. 
The tree-based orders, such as the SYN and BLT orders do not perform well, indicating that predicting words following a syntactic path is not preferable.
On the other hand, Table~\ref{table:D3} shows that the model with SAO achieves competitive and even statistically significant improvements over the L2R order. The improvements are larger for Turkish and Japanese, indicating that a flexible generation order may improve the translation quality for languages with different syntactic structures from English.

\begin{table}[t]
	\centering
	\small
	\scalebox{1.0}{
	\begin{tabular}{l  cc}
	\toprule
	Model & Training (b/s) & Decoding (ms/s) \\
	\midrule
	L2R & 4.21  & 12.3 \\
	SAO ($b=1$) & 1.12  & 12.5 \\
	SAO ($b=8$) & 0.58  & 12.8 \\
	\bottomrule
    \end{tabular}}
    \caption{Comparison of the L2R order with SAO on running time, where b/s is batches per second and ms/s is ms per sentence. All experiments are conducted on $8$ Nvidia V100 GPUs with $2000$ tokens per GPU. 
    We also compare beam sizes of $1$ and $8$ for SAO to search the best orders during training. We report the decoding speed of all three models based on greedy decoding.}
    \label{table:time}
    \vspace{-10pt}
\end{table}
\begin{figure*}[t]
    \centering
    \includegraphics[width=\linewidth]{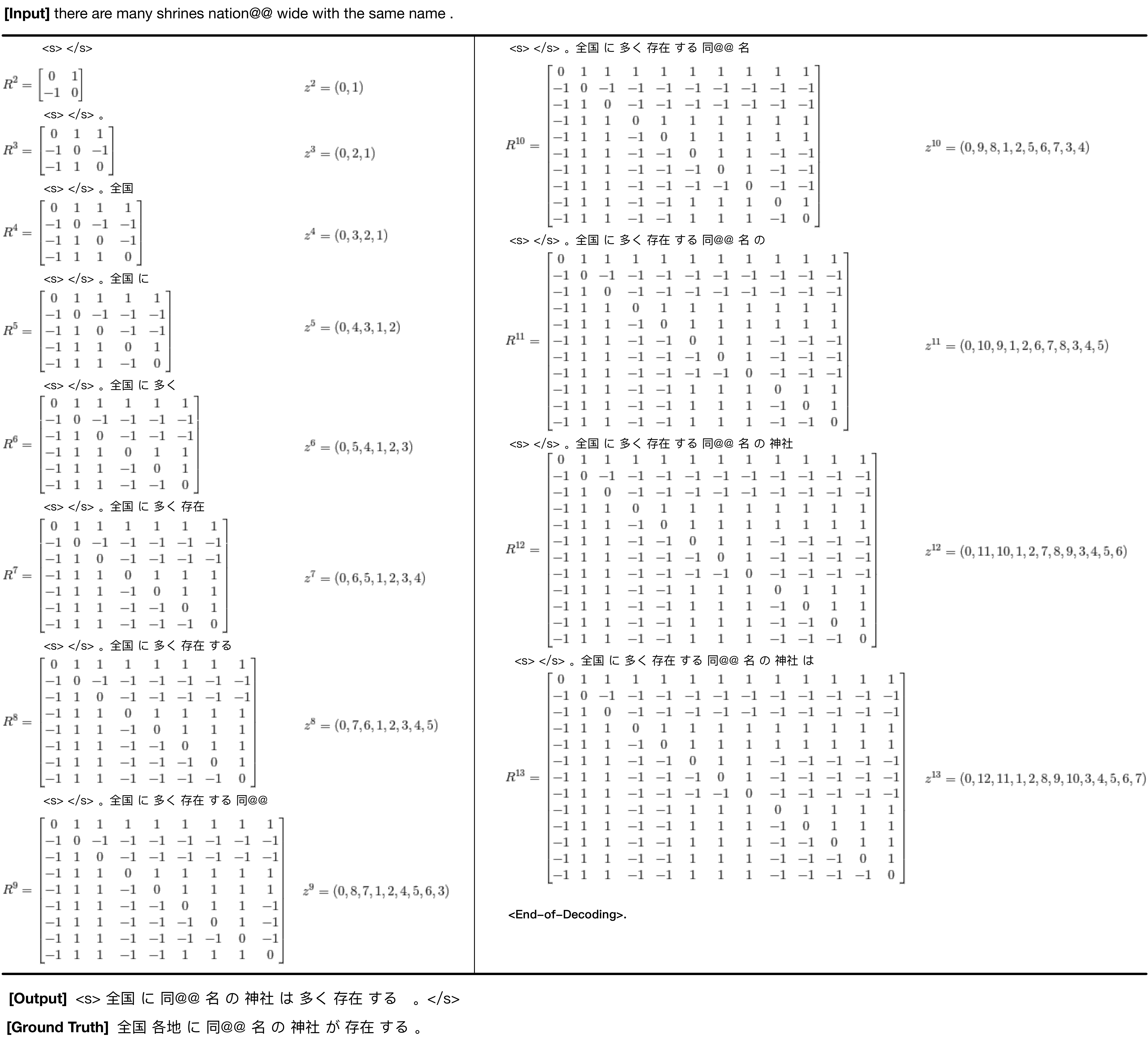}
    \caption{An instantiated concrete example of the decoding process using InDIGO sampled from the En-Ja translation datset. The final output is reordered based on the predicted relative-position matrix.}
    \label{fig:my_label}
\end{figure*}
\begin{figure*}[htbp]
    \centering
    \includegraphics[width=\linewidth]{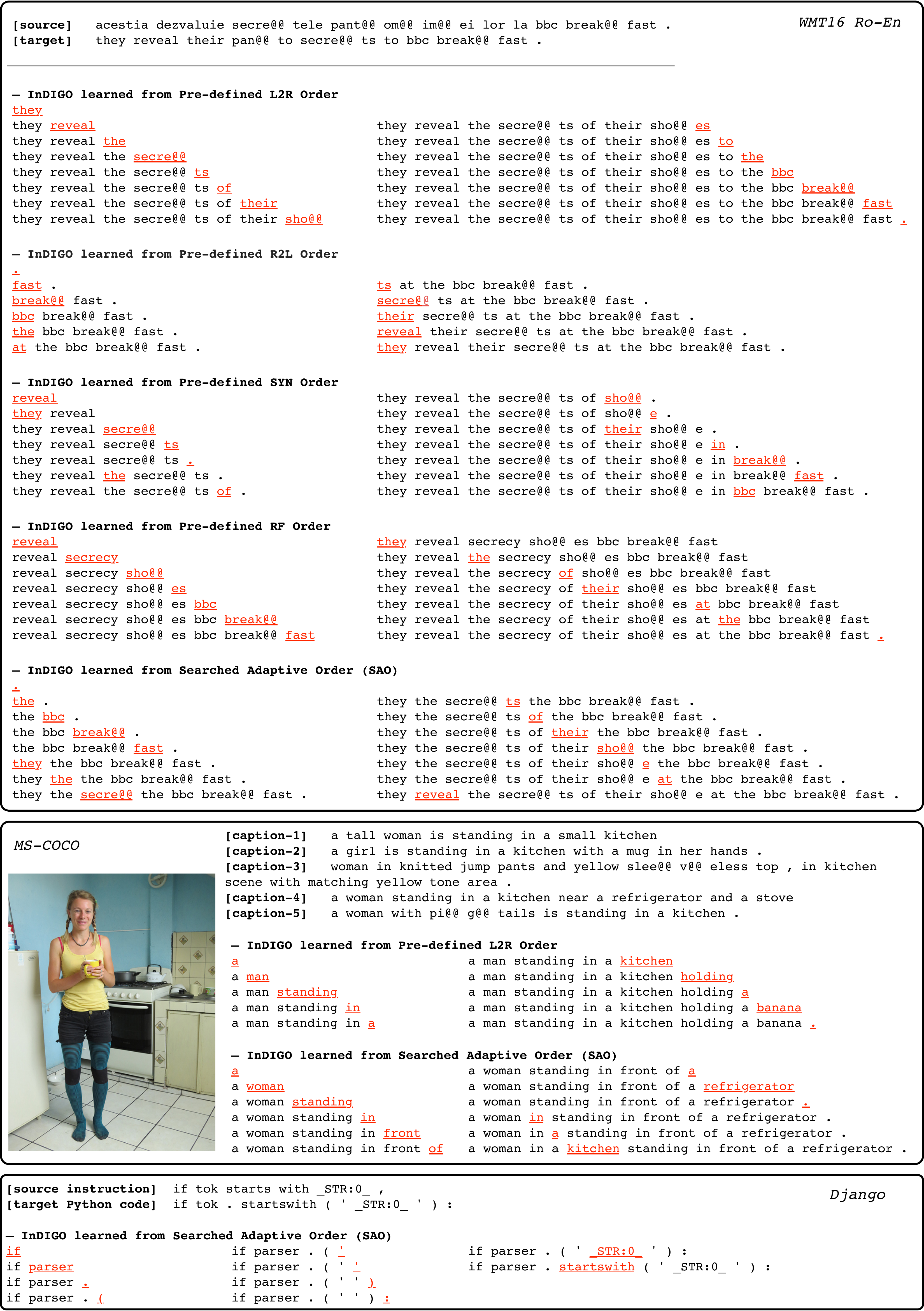}
    \caption{Examples randomly sampled from three tasks that are instructed to decode using InDIGO with various learned generation order. Words in {\color{red}red} and \underline{underlined} are the inserted token at each step. For visually convenience, we reordered all the partial sequences to its correct positions at each decoding step.}
    \label{fig:case_study}
\end{figure*}

\paragraph{Code Generation}

The goal of this task is to generate Python code based on a natural language description, which can be achieved by using a standard sequence-to-sequence generation framework such as the proposed Transformer-InDIGO.
As shown in Table~\ref{table:D4}, SAO works significantly better than the L2R order in terms of both BLEU and accuracy. This shows that flexible generation orders are more preferable in code generation.

\paragraph{Image Captioning}

For the captioning task, one caption is generated per image and is compared against five human-created captions during testing. As show in Table~\ref{table:D4}, we observe that SAO obtains higher BLEU and CIDEr-D~\cite{vedantam2015cider} compared to the L2R order, and it implies that better captions are generated with different orders.

\subsection{Ablation Study}
\paragraph{Model Variants}
Table~\ref{table:ABS} shows the results of the ablation studies using the machine translation task. SAO without bootstrapping nor beam-search degenerates by approximate $1$ BLEU score on Ro-En, demonstrating the effectiveness of these two methods. 
We also test SAO by bootstrapping from a model trained with a R2L order as well as a SYN order, which obtains slightly worse yet comparable results compared to bootstrapping from L2R.
This suggests that the SAO algorithm is quite robust with different bootstrapping methods, and L2R bootstrapping performs the best.
In addition, we re-implement a recent work~\cite{stern2019insertion}, which adopts a similar idea of generating sequences through insertion operations for machine translation. We use the best settings of their algorithm, i.e., training with binary-tree/uniform slot-losses and slot-termination, while removing the knowledge distillation for a fair comparison with ours. Our model obtains better performance compared to \citet{stern2019insertion} on WMT16 Ro-En.

\paragraph{Running Time}
As shown in Table \ref{table:time}, InDIGO decodes sentences as efficient as the standard L2R autoregressive models. However, it is slower in terms of training time using SAO as the supervision, as additional efforts are needed to search the generation orders, and it is difficult to parallelize the SAO. SAO with beam sizes $1$ and $8$ are $3.8$ and $7.2$ times slower compared to L2R, respectively. Note that enlarging the beam size during training won't affect the decoding time as searching the best orders only happen in the training time. We will investigate off-line searching methods to speed up SAO training and make InDIGO more scalable in the future.

\subsection{Visualization}

\paragraph{Relative-Position Matrix}
In Figure~\ref{fig:my_label}, we show an instantiated example produced by InDIGO, which is randomly sampled from the validation set of the KFTT En-Ja dataset. The relative-position matrices ($R^t$) and their corresponding absolute positions ($z^t$) are shown at each step. We argue that relative-position matrices are flexible to encode position information, and its append-only property enables InDIGO to reuse previous hidden states. 

\paragraph{Case Study}
We demonstrate how InDIGO works by uniformly sampling examples from the validation sets for machine translation (Ro-En), image captioning and code generation. As shown in Figure~\ref{fig:case_study}, the proposed model generates sequences in different orders based on the order used for learning (either pre-defined or SAO). For instance, the model generates tokens approximately following the dependency parse wheb we used the SYN order for the machine translation task. On the other hand, the model trained using the RF order learns to first produce verbs and nouns first, before filling up the sequence with remaining functional words.


We observe several key characteristics about the inferred orders of SAO by analyzing the model's output for each task: 
(1) For machine translation, the generation order of an output sequence does not deviate too much from L2R. Instead, the sequences are shuffled with chunks, and words within each chunk are generated in a L2R order; 
(2) In the examples of image captioning and code generation, the model
tends to generate most of the words in the L2R order and insert a few words afterward in certain locations.
Moreover, we provide more examples in the appendix.

\section{Related Work}

\paragraph{Decoding for Neural Models}
Neural autoregressive modelling has become one of the most successful approaches for generating sequences \cite{sutskever2011generating,mikolov2012statistical}, which has been widely used in a range of applications, such as machine translation \cite{sutskever2014sequence}, dialogue response generation \cite{vinyals2015neural}, image captioning \cite{karpathy2015deep} and speech recognition \cite{chorowski2015attention}. Another stream of work focuses on generating a sequence of tokens in a non-autoregressive fashion \cite{gu2017non,lee2018deterministic,oord2017parallel}, in which the discrete tokens are generated in parallel. 
Semi-autoregressive modelling \cite{stern2018blockwise,wang2018semi} is a mixture of the two approaches, while largely adhering to left-to-right generation. Our method is different from these approaches as we support flexible generation orders, while decoding autoregressively. 

\paragraph{Non-L2R Orders}

Previous studies on generation order of sequences mostly resort to a fixed set of generation orders. \citet{wu2018beyond} empirically show that R2L generation outperforms its L2R counterpart in a few tasks. \citet{ford2018importance} devises a two-pass approach that produces partially-filled sentence ``templates" and then fills in missing tokens.
\citet{2019arXiv190100158Z} also proposes to generate tokens by first predicting a text template and infill the sentence afterwards while in a more general way.
\citet{NIPS2018_7796} proposes a middle-out decoder that firstly predicts a middle-word and simultaneously expands the sequence in both directions afterwards. Previous studies also focused on decoding in a bidirectional fashion such as~\cite{sun2017bidirectional,zhou2019synchronous,zhou2019sequence}.
Another line of work models sequence generation based on syntax structures~\cite{yamada2001syntax,charniak2003syntax,chiang2005hierarchical, emami2005neural,zhang2015top,dyer2016recurrent,aharoni2017towards,wang2018tree,eriguchi2017learning}. In contrast, Transformer-InDIGO supports fully flexible generation orders during decoding.

There are two concurrent work ~\cite{welleck2019non,stern2019insertion}, which study sequence generation in a non-L2R order. \newcite{welleck2019non} propose a tree-like generation algorithm. Unlike this work, the tree-based generation order only produces a subset of all possible generation orders compared to our insertion-based models. Further, \newcite{welleck2019non} find L2R is superior to their learned
orders on machine translation tasks, while transformer-InDIGO with searched adaptive orders achieves better performance. 
\newcite{stern2019insertion} propose a very similar idea of using insertion operations in Transformer for machine translation. The major difference is that they directly use absolute positions, while ours utilizes relative positions. As a result, their model needs to re-encode the partial sequence at every step, which is computationally more expensive. In contrast, our approach does not necessitate re-encoding the entire sentence during generation. 
In addition, knowledge distillation was necessary to achieve good performance in \newcite{stern2019insertion}, while our model is able to match the performance of L2R even without bootstrapping.

\section{Conclusion}

We have presented a novel approach -- InDIGO -- which supports flexible sequence generation. Our model was trained with either pre-defined orders or searched adaptive orders. In contrast to conventional neural autoregressive models which often generate from left to right, our model can flexibly generate a sequence following an arbitrary order. Experiments show that our method achieved competitive or even better performance compared to the conventional left-to-right generation on four tasks, including machine translation, word order recovery, code generation and image captioning. 

For future work, it is worth exploring a trainable inference model to directly predict the permutation~\cite{mena2018learning} instead of beam-search. Also, the proposed InDIGO could be extended for post-editing tasks such as automatic post-editing for machine translation (APE) and grammatical error correction (GEC) by introducing additional operations such as ``deletion'' and ``substitution''.

\section*{Acknowledgement}
We specially thank our action editor Alexandra Birch and all the reviewers for their great efforts to review the draft. 
We also would like to thank Douwe Kiela, Marc'Aurelio Ranzato, Jake Zhao and our colleagues at FAIR for the valuable feedback, discussions and technical assistance.
This work was partly supported by
Samsung Advanced Institute of Technology (Next
Generation Deep Learning: from pattern recognition to AI) and Samsung Electronics (Improving
Deep Learning using Latent Structure). KC thanks for the
support of eBay and Nvidia.

\bibliography{acl2019}
\bibliographystyle{acl_natbib}


\end{document}